\documentclass[letterpaper]{article} 
\usepackage[preprint]{aaai2027}
\usepackage[hyphens]{url}  
\usepackage{graphicx} 
\usepackage{natbib}  
\usepackage{caption} 
\usepackage{algorithm}
\usepackage{algorithmic}
\usepackage{amsmath}
\usepackage{amssymb}
\usepackage{mathtools}

\usepackage{booktabs}

\title{How to Guide LLM Generation: Dual-Surrogate Guided Search for \\Automated Heuristic Design}

\author{
  Yuhan Wang\textsuperscript{1}, Chaoda Peng\textsuperscript{1}, Xingyu Wu\textsuperscript{2}, Sheng-Hao Wu\textsuperscript{1}, Zhi-Hui Zhan\textsuperscript{3}
}
\affiliations{
  \textsuperscript{1}College of Mathematics and Informatics, South China Agricultural University\\
  \textsuperscript{2}Department of Data Science and Artificial Intelligence, The Hong Kong Polytechnic University\\
  \textsuperscript{3}College of Artificial Intelligence, Nankai University
}

\newcommand{\method}{DGS}
\newcommand{\fullmethod}{Dual-Surrogate Guided Search}

\begin{document}

\maketitle

\begin{abstract}
Large language models (LLMs) have made automated heuristic design (AHD) increasingly practical by generating executable heuristic code from task descriptions and evaluator feedback. Yet under a limited query and evaluation budget, search efficiency depends critically on a pre-generation decision. Before each LLM query and black-box evaluation, the system must choose which archived heuristics to reuse as parents and which generation operator should transform them. Existing methods typically choose such actions with predefined rules, leaving the expected outcome of each concrete operator-parent action only indirectly modeled. Therefore, we propose \emph{\fullmethod{}} (\method{}), a surrogate-guided action-selection module for operator-parent selection in LLM-based AHD. \method{} guides the LLM code-generation process by scoring pre-generation actions with two complementary surrogates. Specifically, a transition surrogate is proposed to predict the latent distribution of the child representation induced by an operator-parent action, while an instance-conditioned utility surrogate is proposed to estimate the expected performance of sampled child latents. Moreover, we propose an uncertainty-aware acquisition rule that combines predicted utility, utility uncertainty, and transition uncertainty to select the next LLM generation action. Across a diverse heuristic-design suite, \method{} is competitive with strong LLM-AHD baselines, and ablation and action-selection analyses suggest that its behavior goes beyond simple archive ranking or fixed operator preferences.
\end{abstract}


\section{Introduction}

Heuristic algorithms are central to solving difficult optimization problems because they often provide useful solutions when exact optimization is computationally infeasible. However, designing such heuristics is still largely an expert-driven and task-specific process. This trial-and-error workflow is costly and difficult to transfer across tasks, motivating automated heuristic design (AHD): the search for heuristic programs with limited human intervention \citep{zhang2020evolving}. Traditional AHD methods typically search within constrained design spaces, such as symbolic expressions \citep{burke2019classification} or neural networks \citep{luo2023neural,liu2023good}. The rapid progress of code-capable large language models (LLMs) changes this setting: executable heuristic programs can be generated from task specifications and further improved according to previous code and evaluation feedback \citep{wu2024evolutionary,novikov2025alphaevolve,lehman2023evolution}. This capability has given rise to LLM-based AHD \citep{liu2026systematic,kiet2026motif}, where the LLM acts as a program generator inside an automated search process. 

\begin{figure*}[t]
  \centering
  \includegraphics[width=0.98\textwidth]{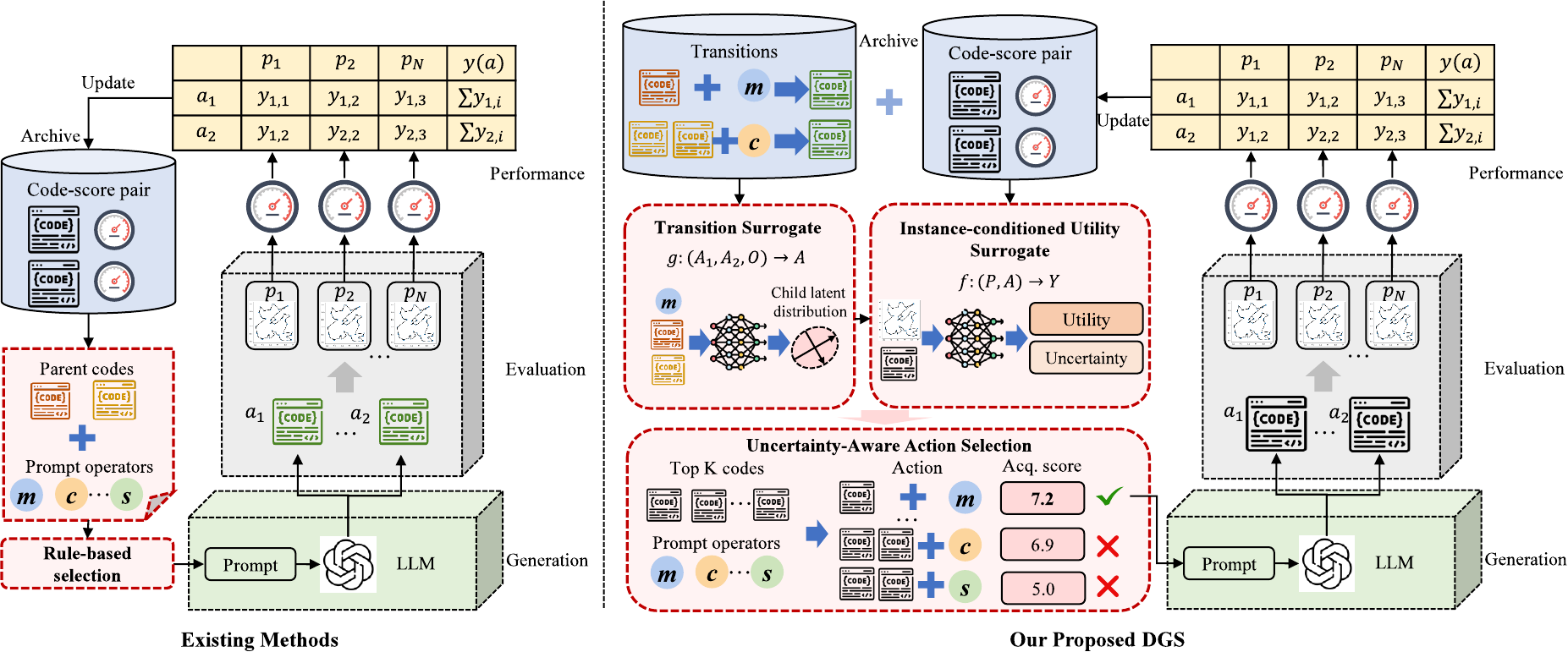}
  \caption{Overview of existing methods and proposed \fullmethod{} (\method{}). A generic archive-based loop reuses archived parent code and generation operators to prompt the LLM. \method{} additionally trains a transition surrogate from parent--operator--child records and an instance-conditioned utility surrogate from per-instance code--score records. The acquisition rule combines predicted utility and uncertainty to choose the next operator-parent action before spending the next generation and evaluation step.}
  \label{fig:teaser}
\end{figure*}

Existing LLM-based AHD methods differ mainly in how this search process is organized. Some methods refine one or a few programs through iterative improvement \citep{van2024llamea}, while population-based methods \citep{liu2024evolution} maintain an archive of candidate heuristics and select promising ones for further generation. Tree-search methods \cite{zheng2025mctsahd} further organize generated heuristics in an explicit search tree to broaden exploration and avoid prematurely committing to a small population. Alongside these search frameworks, prompt design \citep{chen2025hifo} has become a central mechanism for steering heuristic generation. Early methods often prompt the LLM to imitate variation operators from evolutionary algorithms \citep{dat2025hsevo,wu2023transferable}, such as mutation and crossover over existing heuristic code \citep{liu2024evolution}. Later methods enrich this process with reflective feedback \citep{ye2024reevo}, textual gradients \citep{yuksekgonul2025optimizing}, or other forms of natural-language guidance \citep{shi2026generalizable} that explain how previous programs should be improved. These methods have shown strong empirical performance across a range of combinatorial optimization and algorithm-design tasks \citep{zhao2026trajevo}, including routing \citep{hottung2025vrpagent}, packing \citep{sun2026co}, scheduling \citep{yu2026automated}, and black-box optimization \citep{xie2025co,huang2025autonomous,ma2025toward}. Together, these methods establish a common archive--generation--evaluation loop, in which evaluated heuristic programs provide the context for subsequent LLM-based code generations.

Once LLMs can generate executable heuristics, the limiting factor shifts from whether a candidate can be produced to which generation action should be tried next. We study the common black-box-query setting, where each LLM query incurs cost and returns only the generated code, and each evaluator call measures the resulting heuristic through a black-box algorithmic evaluation process. Before spending this budget, the system must choose which archived heuristics to reuse as parents and which prompt operator should transform them. In evolutionary LLM-AHD \citep{liu2024evolution}, this choice is naturally represented as an operator-parent action. Selecting this action is therefore a central way to allocate expensive generation and evaluation steps.

Current systems usually choose these actions with hand-specified search rules, such as archive ranking, stochastic parent selection, tree policies, or fixed prompt-operator schedules. These mechanisms organize the archive and guide generation, but they often do not explicitly learn the expected effect of a concrete operator-parent action from accumulated code-score pairs and parent-child transition records. Consequently, they provide only indirect evidence for deciding which specific operator-parent action should receive the next query and evaluation. This leaves an unresolved action-selection question: how can past evaluations be converted into guidance for the next LLM code-generation query?

Learning such guidance is nontrivial because the object to be selected is not an already instantiated program. It is a pre-generation operator-parent action that determines how the LLM will construct a child heuristic. Before the LLM query, the corresponding child code is unavailable, so the system cannot directly apply a surrogate model to a concrete candidate program as in ordinary surrogate-assisted optimization. Therefore, we propose \emph{\fullmethod{}} (\method{}), a surrogate-guided action-selection module that scores pre-generation actions by estimating both their induced child representations and their expected utility. As summarized in Figure~\ref{fig:teaser}, \method{} combines a transition surrogate, an instance-conditioned utility surrogate, and an uncertainty-aware acquisition rule to select the next generation action before invoking the LLM. The contributions of this paper are:
\begin{itemize}
  \item We formulate pre-generation operator-parent selection as a learned action-selection problem for guiding LLM code generation in AHD.
  \item We propose \fullmethod{}, a dual-surrogate action-selection module that scores pre-generation actions by predicting both their likely child latent distributions and expected utilities.
  \item We introduce an uncertainty-aware acquisition rule that combines predicted utility, utility uncertainty, and transition uncertainty to allocate expensive LLM and evaluator calls.
  \item We provide empirical comparisons, ablation studies, and action-selection analyses that evaluate learned guidance across diverse optimization tasks.
\end{itemize}

\section{Preliminaries}

\subsection{LLM-Based Automated Heuristic Design}

Automated heuristic design (AHD) searches for heuristic programs, components, or selection rules with limited human intervention. Let $\mathcal{P}_{\mathrm{tr}}=\{p_j\}_{j=1}^m$ be the training instances. We view the LLM as a stochastic generator $G$ that maps a prompt $\pi$ to executable code, $a\sim G(\pi)$. We view evaluation as a black-box function $E$ that maps a program and the training set to per-instance scores, $\mathbf{y}(a)=E(a,\mathcal{P}_{\mathrm{tr}})=(y_1(a),\ldots,y_m(a))$. Scores are converted so that larger values are better. The archive score is $y(a)=\sum_{j=1}^m y_j(a)$. Given a finite budget $T$ of generated and evaluated programs, AHD aims to maximize empirical performance on the training instances:
\begin{equation}
  a^\star \in \arg\max_{a\in\mathcal{C}_T}  \frac{1}{m} \sum_{j=1}^m y_j(a),
\end{equation}
where $\mathcal{C}_T$ is the set of valid executable programs generated and evaluated within the budget.

LLM-based AHD implements this search as an archive--generation--evaluation loop. At iteration $t$, the archive $\mathcal{A}_t=\{(a_i,y_i,\mathbf{y}_i,o_i,\mathcal{S}_i)\}_{i=1}^{n_t}$ stores generated code, aggregate and per-instance scores, the operator used to generate the code, and the corresponding parent-index set. Here $y_i=y(a_i)$, $\mathbf{y}_i=\mathbf{y}(a_i)$,  $\mathcal{S}_i\subseteq\{1,\ldots,i-1\}$ indexes the archived programs used as parents, and $\mathcal{S}_i=\emptyset$ for an independent initialization. A new generation is specified by an operator-parent action $x=(o,\mathcal{S})$, where $o$ chooses a prompt operator and $\mathcal{S}\subseteq\{1,\ldots,n_t\}$ indexes archived parent programs. The assembled prompt $\pi(D,o,\mathcal{S})$ combines the task description and function interface $D$, the operator instruction, and the code snippets $\{a_r:r\in\mathcal{S}\}$. The LLM then samples child code
\begin{equation}
  a_{t+1}\sim G(\pi(D,o,\mathcal{S})),
\end{equation}
which is evaluated as $\mathbf{y}_{t+1}=E(a_{t+1},\mathcal{P}_{\mathrm{tr}})$ and appended to the archive together with its aggregate score $y_{t+1}$.

Existing LLM-based AHD methods mainly differ in how they organize $\mathcal{A}_t$ and choose $x$. FunSearch uses a score-based archive to preserve and sample promising programs \citep{romera2024mathematical}. Evolution of Heuristics (EoH) introduces a broadly used evolutionary prompt-operator template over heuristic thoughts and code, covering initialization, mutation-style transformation, and recombination-style generation \citep{liu2024evolution}. ReEvo adds reflective feedback and verbal gradients to describe how previous programs should be improved \citep{ye2024reevo}. Monte Carlo Tree Search for AHD (MCTS-AHD) stores generated heuristics in a tree and uses tree policies to select expansion directions. It also builds on the EoH-style operator set by adding one prompt operator \citep{zheng2025mctsahd}. Despite these advances, action choice remains largely rule-based or policy-driven and does not explicitly model the expected outcome of each concrete operator-parent action.

Recent work further improves the generation process through mutation-control prompts, hindsight/foresight or metacognitive prompting, LLM co-evolution, structural feedback, coding-agent evolution, and prompt-strategy optimization \citep{yin2024controlling,yao2024multi,huang2025calm,vanstein2026llamea,novikov2025alphaevolve}. Since both LLM queries and heuristic quality evaluations are black-box and expensive, LLM-based AHD should use the archive to guide generation sample-efficiently. This motivates learning action-level guidance from code-score records and parent-child transitions, so that the next operator-parent action is selected using evidence accumulated during the search.

\subsection{Surrogate-Guided Search}

Bayesian optimization and surrogate-assisted search use past evaluations to fit a cheap predictive model and choose subsequent evaluations through an acquisition rule \citep{srinivas2009gaussian,wang2023recent}. A common example is the Gaussian-process upper confidence bound (GP-UCB), which selects candidates by combining predicted value and predictive uncertainty \citep{srinivas2009gaussian}. In standard black-box optimization, the surrogate scores candidate solutions or parameter settings that already exist. LLM-based AHD changes this decision point: before the LLM is queried, the child code associated with an operator-parent action is unavailable. \method{} therefore uses a transition surrogate to map a pre-generation action to a predicted child-latent distribution, and then uses an instance-conditioned utility surrogate to estimate the value of sampled child latents. This formulation turns archive data into learned pre-generation guidance rather than only ranked memory or post-generation candidate scoring.

\section{\fullmethod{}}
\label{sec:method}

\subsection{Method Overview}

\method{} guides LLM code generation by selecting an operator-parent action before each expensive generation call. Given the current archive $\mathcal{A}_t$, it chooses $x=(o,\mathcal{S})$ and converts it into an operator-specific prompt. Because $x$ is a pre-generation instruction, its child code and score are observed only after the LLM and evaluator are called. \method{} reuses five fixed EoH-style prompt operators \citep{liu2024evolution}, $\mathcal{O}=\{\texttt{i1},\texttt{m1},\texttt{m2},\texttt{e1},\texttt{e2}\}$. The \texttt{i1} operator generates initial programs from the task description and function template. Given one parent, \texttt{m1} asks the LLM to introduce new mechanisms or code segments, while \texttt{m2} varies important parameters or equations. Given two parents, \texttt{e1} generates a structurally different algorithm inspired by them, while \texttt{e2} mimics crossover in evolutionary algorithms by combining information from both parents. \method{} learns which concrete operator-parent action to use while keeping these prompt templates fixed. For detailed description of the five operators, refer to the supplementary material. Note that \method{} is compatible with other prompt operator set $\mathcal{O}$. We also conduct experiments to verify the effectiveness of DGS for different operator sets.

\method{} learns from two archive-derived datasets. Code-score records contain generated code and its aggregate and per-instance scores on $\mathcal{P}_{\mathrm{tr}}$ and are used to train the instance-conditioned utility surrogate. Transition records link each executed operator-parent action to its child and are used to train the transition surrogate. As summarized in Algorithm~\ref{alg:dgs}, the search initializes $\mathcal{A}$ with $n_0$ independent \texttt{i1} generations, samples $n_w$ random actions to collect initial transition data, and then begins guided search. At each guided iteration, \method{} updates the surrogates from $\mathcal{A}_t$, forms the candidate parent-index set $\mathcal{B}_t\subseteq\{1,\ldots,n_t\}$, and enumerates its feasible actions. To control online training cost, it alternates periodic full updates of the shared representation and both surrogates with fast intermediate updates. The latter freeze the representation and update only the utility heads and transition surrogate. The action with the largest acquisition value is sent to the LLM, and its generated child is evaluated and appended to the archive.

\subsection{Shared Latent Representation}

\method{} represents heuristic code, problem instances, and operators in learned continuous spaces. A frozen code/text encoder first maps the code of heuristic $a$ to a code embedding $e(a)$. We use ModernBERT for this encoder because of its long-context encoding capability \citep{warner2025smarter}. The method can also use other code/text encoders. The embedding provides a semantic representation of the code, but it is not trained to reflect the performance landscape of the target AHD task. We therefore learn a lightweight task-adaptive representation that maps $e(a)$ into a latent representation \citep{maus2022local,ahmed2026latent}:
\begin{equation}
  z(a)=\mathrm{norm}\left(W e(a) + \rho R_\phi(e(a))\right)
  \in\mathbb{R}^{d_z},
\end{equation}
where $W$ is a skip projection, $R_\phi$ is a residual multilayer perceptron (MLP), $\rho$ controls the residual scale, and $d_z$ denotes the latent representation dimension. The normalization keeps all heuristic latents on a comparable scale. Each training instance $p_j$ is assigned a learnable instance embedding $q_j\in\mathbb{R}^{d_z}$, indexed by its identifier. We adopt this task-agnostic choice to preserve the generality of the proposed method. Note that more advanced task-specific instance embeddings can also be employed, such as Transformer-based instance embeddings for routing problems \citep{ma2021learning}. The instance embedding dimensionality is kept the same as that of the heuristic latent representation $z(a)$ to enable the interaction-feature computation defined below. For an operator $o\in\mathcal{O}$, $v_o=\eta_\omega(o)\in\mathbb{R}^{d_o}$ is a learned lookup embedding that is trained jointly with the transition surrogate to distinguish mutation-like and recombination-like actions.

\subsection{Instance-Conditioned Utility Surrogate}

The instance-conditioned utility surrogate predicts the performance of candidate child latents on individual training instances rather than reducing each heuristic to one aggregate archive score. Its design combines an interaction representation that couples heuristic and instance latents, an ensemble that estimates predictive mean and uncertainty, and a training objective that fits normalized scores while preserving within-instance rankings.

Inspired by \citet{wu2024large}, we propose to augment the utility-head input with additional heuristic--instance interaction features. Given a heuristic latent $z$ and an instance embedding $q_j$, we define
\begin{equation}
  \xi(z,q_j) = [z,\ q_j,\ z\odot q_j,\ |z-q_j|,\ \cos(z,q_j)].
\end{equation}
The additional terms $z\odot q_j$, $|z-q_j|$, and $\cos(z,q_j)$ model complementary forms of heuristic--instance compatibility. An ensemble \citep{zhou2002ensembling} of $H$ utility heads maps this feature to per-instance utility predictions:
\begin{equation}
  \hat{u}^{(h)}_j(z)=f_h(\xi(z,q_j)), \qquad h=1,\ldots,H.
\end{equation}
For each instance, $\bar{u}_j(z)=\operatorname{Mean}_{h=1}^H[\hat{u}^{(h)}_j(z)]$ and $s_{u,j}(z)=\operatorname{Std}_{h=1}^H[\hat{u}^{(h)}_j(z)]$ summarize the ensemble mean and disagreement. Summing them over training instances gives $\mu_u(z)=\sum_{j=1}^m \bar{u}_j(z)$ and $\sigma_u(z)=\sum_{j=1}^m s_{u,j}(z)$, respectively. To diversify the heads, each head independently draws $n_t$ archived algorithms with replacement to construct the training table. Every draw contributes all rows of that algorithm across training instances. The utility training table contains one row for each algorithm--instance pair:
\begin{equation}
  \mathcal{T}_u=\{(e(a_i), j, \bar{y}_{ij}, i): i=1,\ldots,n_t,\ j=1,\ldots,m\},
\end{equation}
where the last index identifies the source algorithm. The target $\bar{y}_{ij}$ equals the recorded score $y_{ij}$ when it is valid. Otherwise, it is set to $\bar{y}_j-4s_j$, where $\bar{y}_j$ and $s_j$ are the mean and standard deviation of valid scores on instance $p_j$. All targets are standardized over $\mathcal{T}_u$. Pointwise MSE fits normalized score magnitudes, but action selection ultimately depends on the ordering of candidate heuristics. \method{} therefore adds within-instance rank supervision to preserve pairwise preferences without comparing scores across heterogeneous instances. For rank supervision \citep{tan2025offline,ma2025surrogate}, $(i,j)\in\mathcal{M}$ means that the row associated with algorithm $i$ and instance $j$ is present in mini-batch $\mathcal{M}\subset\mathcal{T}_u$. The preference pairs are
\begin{equation}
  \mathcal{R}^{+}
  =
  \{(i,i',j):(i,j),(i',j)\in\mathcal{M},\ \tilde{y}_{ij}>\tilde{y}_{i'j}\},
\end{equation}
where $\tilde{y}_{ij}$ denotes the standardized target. For a given head, write $\hat{u}_{ij}=\hat{u}^{(h)}_j(z(a_i))$, $\Delta_{ii'j}=\hat{u}_{ij}-\hat{u}_{i'j}$, and $\sigma_s(v)=(1+e^{-v})^{-1}$. The rank loss is
\begin{equation}
  \mathcal{L}_{\mathrm{rank}}=-|\mathcal{R}^{+}|^{-1}\sum\nolimits_{(i,i',j)\in\mathcal{R}^{+}}\log\sigma_s(\Delta_{ii'j}).
\end{equation}
If a batch contains no valid within-instance pair, the rank term is set to zero. The per-instance mean-squared-error (MSE) loss for head $h$ is
\begin{equation}
  \mathcal{L}^{(h)}_{\mathrm{mse}}(\mathcal{M})
  =
  |\mathcal{M}|^{-1}
  \sum_{(i,j)\in\mathcal{M}}
  (\hat{u}^{(h)}_j(z(a_i))-\tilde{y}_{ij})^2 .
\end{equation}
During a full surrogate update, the representation is trained together with one anchor utility head, denoted by $h=h_{\mathrm{anc}}$. Let $z_i^{-}$ be the cached latent representation of $a_i$ before the representation update. It is detached and treated as a fixed target with no gradient. The drift penalty discourages abrupt changes in the latent geometry during representation updates:
\begin{equation}
  \mathcal{L}_{\mathrm{drift}}(\mathcal{M})
  =
  |\mathcal{M}|^{-1}
  \sum_{(i,j)\in\mathcal{M}}
  \|z(a_i)-z_i^{-}\|_2^2 .
\end{equation}
The full-update loss is
\begin{equation}
  \mathcal{L}_{\mathrm{full}}
  =
  \mathcal{L}^{(h_{\mathrm{anc}})}_{\mathrm{mse}}(\mathcal{M})
  +\lambda_r\mathcal{L}_{\mathrm{rank}}(\mathcal{M})
  +\lambda_d\mathcal{L}_{\mathrm{drift}}(\mathcal{M}) .
\end{equation}
After this representation update, the representation is frozen. Each utility head is then trained on its independently drawn algorithm-level bootstrap sample. For head $h$, the loss is
\begin{equation}
  \mathcal{L}^{(h)}_{\mathrm{head}}
  =
  \mathcal{L}^{(h)}_{\mathrm{mse}}(\mathcal{M}_h)
  +\lambda_r\mathcal{L}_{\mathrm{rank}}(\mathcal{M}_h).
\end{equation}

\subsection{Transition Surrogate}

The transition surrogate is needed because an operator-parent action does not directly provide a candidate program to score. A candidate action $x=(o,\mathcal{S})$ specifies an operator $o$ and a parent set $\mathcal{S}$, while the child code is observed only after querying the LLM. Therefore \method{} learns an action-to-latent transition model from archived transition records.

A transition training tuple consists of an executed action and the resulting child sample identifier. The transition table is
\begin{equation}
  \mathcal{T}_\tau=\{((o_i,\mathcal{S}_i),c_i)\}_{i=1}^{N_\tau},
\end{equation}
where $N_\tau$ is the number of executed transitions, $\mathcal{S}_i$ is the parent-index set, and $c_i$ is the archive index of the child produced by operator $o_i$ from those parents. The parent and child code are encoded through the shared representation. For an action $x=(o,\mathcal{S})$, the transition input contains a distributional summary of the parent latents and the jointly learned operator embedding $v_o$. Let $\{z(a_r):r\in\mathcal{S}\}$ be the parent latents and let $\epsilon$ be a small base variance. The parent distribution is
\begin{equation}
  \mu_{\mathrm{in}}(\mathcal{S})
  =
  \frac{1}{|\mathcal{S}|}\sum_{r\in\mathcal{S}} z(a_r),
\end{equation}
\begin{equation}
  \sigma^2_{\mathrm{in}}(\mathcal{S})
  =
  \begin{cases}
  \epsilon\mathbf{1}, & |\mathcal{S}|=1,\\
  \frac{1}{|\mathcal{S}|}\sum_{r\in\mathcal{S}}
  (z(a_r)-\mu_{\mathrm{in}}(\mathcal{S}))^2+\epsilon\mathbf{1}, & |\mathcal{S}|>1.
  \end{cases}
\end{equation}
The transition MLP receives the concatenated input $[\mu_{\mathrm{in}}(\mathcal{S}),\log\boldsymbol{\sigma}^2_{\mathrm{in}}(\mathcal{S}),v_o]\in\mathbb{R}^{2d_z+d_o}$, where the first two components lie in $\mathbb{R}^{d_z}$ and $v_o\in\mathbb{R}^{d_o}$. It predicts the mean and log-variance of a diagonal Gaussian distribution over the child latent:
\begin{equation}
  (\mu_\tau(x), \log \boldsymbol{\sigma}_\tau^2(x))
  =
  g_\psi([\mu_{\mathrm{in}}(\mathcal{S}),\log\boldsymbol{\sigma}^2_{\mathrm{in}}(\mathcal{S}),v_o]).
\end{equation}
Both $\mu_\tau(x)$ and $\log \boldsymbol{\sigma}_\tau^2(x)$ lie in $\mathbb{R}^{d_z}$. The transition map $g_\psi$ parameterizes residual updates with respect to the parent distribution, followed by latent normalization and log-variance clipping. We denote the resulting child-latent distribution by $\tau(x)$. For a transition mini-batch $\mathcal{M}_\tau\subset\mathcal{T}_\tau$, the transition head is trained against the observed child latent $z(a_c)$. Let $b_\tau=|\mathcal{M}_\tau|$, $\Sigma_x=\operatorname{diag}(\boldsymbol{\sigma}_\tau^2(x))$, $\delta_{x,c}=z(a_c)-\mu_\tau(x)$, and $\|v\|_A^2=v^\top A v$. The Gaussian negative log-likelihood is
\begin{equation}
  \mathcal{L}_{\tau}=\frac{1}{2b_\tau}\sum_{(x,c)\in\mathcal{M}_\tau}\left[\log|\Sigma_x|+\|\delta_{x,c}\|_{\Sigma_x^{-1}}^2\right].
\end{equation}
This objective omits only the constant term of the Gaussian likelihood. Unlike the utility surrogate, which evaluates a realized or sampled latent representation, the transition surrogate does not predict utility directly. It predicts the child-latent distribution induced by a pre-generation action. We compute its scalar uncertainty as $s_\tau(x)=\|\boldsymbol{\sigma}_\tau(x)\|_2$, which is used below in uncertainty-aware action selection.

\subsection{Uncertainty-Aware Action Selection}

At each guided iteration, \method{} follows the population construction used in EoH and MCTS-AHD: valid archive entries are ranked by aggregate score and the indices of the top-$K$ entries form $\mathcal{B}_t\subseteq\{1,\ldots,n_t\}$. Thus parent selection is not performed over the entire archive. \method{} instead applies surrogate-guided selection to the operator-parent actions induced by this filtered population:
\begin{equation}
  \mathcal{X}_t=\{(o,\mathcal{S}):o\in\mathcal{O}_{\mathrm{sel}},\ \mathcal{S}\subseteq\mathcal{B}_t,\ |\mathcal{S}|=\kappa(o)\},
\end{equation}
where $\mathcal{O}_{\mathrm{sel}}=\{\texttt{m1},\texttt{m2},\texttt{e1},\texttt{e2}\}$ during guided selection and $\kappa(o)$ is the parent arity of operator $o$. The operators \texttt{m1} and \texttt{m2} each use one parent, while \texttt{e1} and \texttt{e2} each use two parents. For a parent pool of size $K$, these arities yield at most $2K+2\binom{K}{2}$ feasible actions.

Given the child-latent distribution $\tau(x)$ predicted by the transition surrogate, \method{} defines the acquisition score of an action $x$ as
\begin{equation}
  \begin{aligned}
  \alpha(x)
  =&\ \mathrm{E}_{\tilde{z}\sim \tau(x)}[\mu_u(\tilde{z})]
  + \beta_u \mathrm{E}_{\tilde{z}\sim \tau(x)}[\sigma_u(\tilde{z})] \\
  &+ \beta_u \lambda_\tau |\mathcal{P}_{\mathrm{tr}}| s_\tau(x).
  \end{aligned}
  \label{eq:acq}
\end{equation}
The first term exploits actions predicted to produce high-utility children. The second term encourages exploration where the utility ensemble disagrees. The third term encourages exploration where the transition surrogate is uncertain about the generation outcome. Here $\beta_u$ controls the overall exploration strength, while $\lambda_\tau$ controls the relative contribution of transition uncertainty within that exploration bonus. The factor $|\mathcal{P}_{\mathrm{tr}}|$ keeps the transition-uncertainty bonus on the same scale as the instance-summed utility prediction.

In implementation, \method{} estimates the expectations in Equation~\ref{eq:acq} by drawing $M_{\mathrm{MC}}$ child latents $\tilde{z}\sim\tau(x)$ and averaging their utility-ensemble predictions. The action with the largest estimated acquisition value is converted into an operator-specific prompt containing the selected parent code, and the generated child is evaluated and appended to the archive.

\begin{algorithm}[tb]
\caption{\fullmethod{} workflow}
\label{alg:dgs}
\textbf{Input}: $D$, $G$, $E$, $\mathcal{P}_{\mathrm{tr}}$, $T$\\
\textbf{Parameter}: $n_0$, $n_w$, $K$, $M_{\mathrm{MC}}$, $\beta_u$, $\lambda_\tau$\\
\textbf{Output}: best valid heuristic in $\mathcal{A}$
\begin{algorithmic}[1]
\STATE Initialize archive $\mathcal{A}$ with $n_0$ independent \texttt{i1} generations from $G$ and evaluations by $E$
\FOR{$\ell=1,\ldots,n_w$}
  \STATE Form top-$K$ parent pool $\mathcal{B}$ from $\mathcal{A}$ and enumerate feasible actions $\mathcal{X}$
  \STATE Sample a warmup action $x=(o,\mathcal{S})$ from $\mathcal{X}$
  \STATE Generate $a\sim G(\pi(D,o,\mathcal{S}))$ and evaluate $\mathbf{y}(a)=E(a,\mathcal{P}_{\mathrm{tr}})$
  \STATE Append code, scores, operator, parents, and transition record to $\mathcal{A}$
\ENDFOR
\FOR{$t=n_0+n_w+1,\ldots,T$}
  \STATE Build $\mathcal{T}_u$ and $\mathcal{T}_\tau$ from $\mathcal{A}$
  \STATE Train or update the representation, utility ensemble, and transition surrogate
  \STATE Form top-$K$ parent pool $\mathcal{B}_t$ from $\mathcal{A}$ and enumerate $\mathcal{X}_t$
  \FORALL{$x\in\mathcal{X}_t$}
    \STATE Predict $\tau(x)$ with the transition surrogate
    \STATE Estimate $\alpha(x)$ using $M_{\mathrm{MC}}$ sampled child latents, the utility ensemble, and Equation~\ref{eq:acq}
  \ENDFOR
  \STATE Select $x_t=(o_t,\mathcal{S}_t)=\arg\max_{x\in\mathcal{X}_t}\alpha(x)$
  \STATE Generate $a_t\sim G(\pi(D,o_t,\mathcal{S}_t))$ and evaluate $\mathbf{y}(a_t)=E(a_t,\mathcal{P}_{\mathrm{tr}})$
  \STATE Append code, scores, operator, parents, and transition record to $\mathcal{A}$
\ENDFOR
\STATE \textbf{return} best valid heuristic in $\mathcal{A}$
\end{algorithmic}
\end{algorithm}

\section{Experimental Study}
\label{sec:experiments}

\begin{figure*}[t]
\centering
  \includegraphics[width=0.85\textwidth]{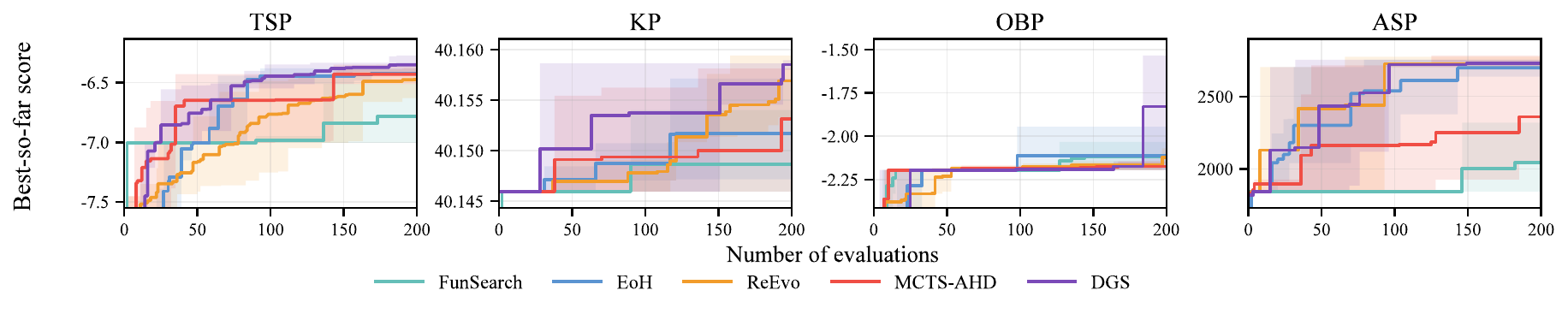}
\caption{Evolution curves on four tasks. Curves show mean best-so-far scores; shaded bands show the min--max range.}
\label{fig:evolution_core}
\end{figure*}

The experiments are organized around three research questions (RQs):
\begin{itemize}
    \item \textbf{RQ1.} Does DGS improve sample efficiency across AHD tasks under limited LLM-query and black-box evaluation budgets?
    \item \textbf{RQ2.} Is learned joint operator-parent selection necessary for the observed gains beyond the shared prompt operators and parent pool?
    \item \textbf{RQ3.} Do the surrogate components contribute to the effectiveness of the proposed action-selection mechanism?
\end{itemize}

\subsection{Experimental Setup}

To answer the first question, we follow common benchmark setups in LLM-based AHD \citep{liu2024evolution,zheng2025mctsahd,liu2024llm4ad} and evaluate \method{} on five heuristic-design tasks. In the Traveling Salesman Problem (TSP) \citep{LKH3}, Knapsack Problem (KP) \citep{jaszkiewicz2002performance}, Online Bin Packing (OBP) \citep{seiden2002online}, and Admissible Set Problem (ASP) \citep{zhang2024understanding}, the generated program is a constructive heuristic that directly builds a solution. In Capacitated Vehicle Routing Problem with Ant-Colony Optimization (CVRP-ACO) \citep{uchoa2017new,ye2024deepaco}, the generated program is a heuristic component that guides a stochastic ant-colony solver. These tasks therefore cover direct constructive heuristics and a solver-embedded heuristic component.

Each search run is capped at 200 generated and evaluated programs, and each method is run three times unless otherwise noted. All methods use \texttt{GPT-4o-mini} as the code-generation LLM. We compare \method{} with four representative LLM-AHD baselines under the same benchmark setting: FunSearch, EoH, ReEvo, and MCTS-AHD \citep{romera2024mathematical,liu2024evolution,ye2024reevo,zheng2025mctsahd,liu2024llm4ad}. These baselines cover archive-based program evolution, evolutionary prompt operators, reflective feedback, and tree-search-based parent selection. For test evaluation, each run first selects the best valid program found on the training instances, and the selected program is then evaluated on the corresponding test setting. Because the tasks use different natural metrics, task-specific results are provided in the supplementary material and average ranks are used in the result summary.

\subsection{Main Results}

Table~\ref{tab:main_summary} summarizes average ranks by task and the complete per-task results are provided in the supplementary material. \method{} obtains the best average rank on TSP, KP, and OBP, ties ReEvo on ASP, and ranks second on CVRP-ACO. The detailed test results are especially strong on TSP, KP, and OBP: \method{} gives the best TSP performance at all reported sizes, obtains the best average rank on KP, and ranks first on every displayed OBP setting. Because \method{} keeps the EoH-style prompt operators and top-$K$ parent-pool construction while replacing rule-based generation choices with dual-surrogate guidance over joint operator-parent actions, these results highlight the value of learning which operator and parents to use for the next LLM query and evaluation.

\begin{table}[t]
\centering
\small
\setlength{\tabcolsep}{1.2mm}
\begin{tabular}{lccccc}
\toprule
Method & TSP & KP & OBP & ASP & CVRP \\
\midrule
FunSearch & 5.00 & 4.50 & 3.50 & 5.00 & 4.67 \\
EoH & 3.00 & 3.50 & 2.42 & 2.33 & 3.00 \\
ReEvo & 3.75 & 2.50 & 3.75 & \textbf{1.83} & 3.33 \\
MCTS-AHD & 2.25 & 3.25 & 4.33 & 4.00 & \textbf{1.67} \\
\method{} & \textbf{1.00} & \textbf{1.25} & \textbf{1.00} & \textbf{1.83} & 2.33 \\
\bottomrule
\end{tabular}
\caption{Method-by-task summary of the main comparison. Entries are average ranks; lower is better. Complete per-task results are provided in the supplementary material.}
\label{tab:main_summary}
\end{table}

To examine search dynamics, we plot best-so-far search trajectories in Figure~\ref{fig:evolution_core}. The curves show how each method converts evaluated programs into improved archive entries during search. After the initial archive is available, surrogate-guided action selection produces faster improvements on several tasks and reaches stronger final archive entries on the main comparison settings. These results address \textbf{RQ1}: learned operator-parent guidance improves the sample efficiency of LLM queries and black-box evaluations by turning past evaluations into more productive subsequent generation actions.

\subsection{Effect of Joint Operator-Parent Selection}

To isolate the effect of action selection from broader framework differences, we construct a controlled TSP comparison in which all variants use the same prompt operators, search budget, and top-$K$ parent-pool construction. The only changed component is how the next operator-parent action is selected. We evaluate two groups of controls. The rule-based controls select parents by archive rank and choose operators by round-robin (H1), random sampling (H2), or an online operator bandit (H3) \citep{auer2002finite}. The surrogate controls use the same archive data as DGS, but restrict the learned score to joint operator-parent action (H4), the operator only (H5), or the parent choice only (H6). These controls rank actions using predicted mean utility only. The results in Table~\ref{tab:controlled_action_selection} show that the joint operator-parent surrogate performs best, improving over the strongest non-surrogate control. The operator-only and parent-only variants are weaker, suggesting that the useful decision variable is the concrete operator-parent action rather than either axis alone. Among these controls, the strongest non-surrogate variant is an operator-bandit rule combined with rank-based parent selection, which already captures a reasonable hand-designed preference over operators. The joint operator-parent surrogate performs better, showing that jointly learning which operator and parents to use improves search beyond fixed selection rules. This controlled comparison supports the \textbf{RQ2} conclusion that the gain comes from learned joint operator-parent selection rather than from the shared prompt operators or top-$K$ parent-pool construction.

\begin{table}[t]
\centering
\small
\setlength{\tabcolsep}{1mm}
\begin{tabular}{clc}
\toprule
Config & Action-selection rule & Mean best \\
\midrule
H1 & round-robin + rank & -6.619 \\
H2 & random + rank & -6.612 \\
H3 & operator bandit + rank & -6.449 \\
H4 & joint operator-parent surrogate & \textbf{-6.406} \\
H5 & operator-only surrogate & -6.802 \\
H6 & parent-only surrogate & -6.701 \\
\bottomrule
\end{tabular}
\caption{Controlled action-selection study on TSP. The metric is the three-run mean best training score; higher is better.}
\label{tab:controlled_action_selection}
\end{table}

\subsection{Ablation Study}

To examine whether the internal components of DGS are necessary, we construct ablation variants by changing one modeling choice at a time while keeping the prompt/operator templates and task settings fixed. Table~\ref{tab:dgs_ablation_train} compares these variants on TSP and OBP using the best score found on the training instances. DGS (Original) obtains the best score on both tasks. Removing interaction features or instance conditioning weakens heuristic--instance utility modeling. Removing the utility ensemble, rank supervision, or drift regularization weakens utility learning and representation updates. Eliminating transition uncertainty or transition prediction weakens action-outcome modeling. We further evaluate the ablation variants on the TSP and OBP test settings, with detailed configuration descriptions and result analysis provided in the supplementary material. Several variants that remain plausible on the training objective lose quality or validity under test evaluation, especially on OBP. This indicates that the surrogate helps the search identify selected programs that remain executable and evaluable beyond the training instances. These ablations support the \textbf{RQ3} conclusion that DGS relies on the coupled utility, transition, and representation-learning components rather than on a single fixed selection rule.

\begin{table}[t]
\normalfont\small
\centering
\begin{tabular}{lrr}
\toprule
Configuration & TSP & OBP \\
\midrule
DGS (Original) & \textbf{-6.351} & \textbf{-1.829} \\
w/o $\mathcal{L}_{\mathrm{rank}}$ & -6.750 & -2.187 \\
w/o $\beta_u$ & -6.464 & -2.174 \\
w/o $\mathcal{L}_{\mathrm{drift}}$ & -6.657 & -2.191 \\
w/o $\lambda_\tau$ & -6.899 & -2.187 \\
w/o utility ensemble & -6.486 & -2.166 \\
w/o $z\odot q, |z-q|$ & -6.548 & -2.195 \\
w/o instance conditioning & -6.582 & -2.195 \\
w/o transition prediction & -6.477 & -2.166 \\
\bottomrule
\end{tabular}
\caption{Ablation comparison for the final DGS configuration. Values are three-run mean best scores on the training instances; higher is better. Each ablation changes only the named component under the same prompt/operator templates.}
\label{tab:dgs_ablation_train}
\end{table}

\subsection{Additional Analysis}

We further conduct three additional analyses, with detailed results provided in the supplementary material. First, we test the learned representation in an offline heuristic selection setting over fixed pregenerated candidate pools. A GP-UCB selector built on the utility-trained latent representations usually reduces simple regret faster than selectors built on frozen code embeddings, supporting the role of utility-supervised latent learning in making code representations more predictive. Second, we analyze operator-usage sequences, raster plots, and sliding-window usage curves during the DGS process. The observed behavior shows adaptive operator preferences across tasks and search stages, indicating that DGS is not merely following a single fixed operator preference. Third, we evaluate DGS with the tree-path synthesis operator introduced by MCTS-AHD \citep{zheng2025mctsahd}. DGS remains effective under the expanded operator vocabulary and achieves competitive or improved performance on several tasks, supporting the applicability of learned action selection across different operator sets.

Overall, \method{} allocates LLM generation and evaluation calls through learned action-level guidance rather than altered prompts. The operator-set analysis further shows that this guidance extends to a broader generation vocabulary, with task-dependent gains.

\section{Conclusion}

This paper studied how an LLM-based automated heuristic design system can turn past evaluations into guidance for the next generation action. We introduced \fullmethod{}, a surrogate-guided action-selection module over operator-parent actions. \method{} combines a transition surrogate, a utility surrogate, and an uncertainty-aware acquisition rule. Together, these components estimate the child-latent distribution induced by a pre-generation action and the expected utility of that induced outcome before spending the next LLM query and evaluation. The empirical results support this action-level view. Learned guidance achieves strong performance against representative LLM-AHD baselines, and the ablation and action-selection analyses suggest that the behavior is not merely copying archive ranks or fixed operator preferences. Future work can further improve uncertainty calibration and extend this form of learned generation guidance to broader AHD settings.

\bibliography{aaai2027}

\end{document}